\documentclass[letterpaper, 10 pt, journal, twoside]{IEEEtran}  

\usepackage{graphicx} 
\graphicspath{{figures/}}

\usepackage{amsthm}
\usepackage{amsmath} 
\usepackage{amssymb}  
\usepackage{mathtools}
\usepackage{ulem} 
\usepackage{multirow}
\usepackage{verbatim}
\usepackage{soul}
\usepackage{algorithm}
\usepackage{algpseudocode}

\usepackage{color}
\usepackage{xcolor}
\usepackage{latexsym}
\usepackage{multicol}
\usepackage{booktabs}
\usepackage{adjustbox}
\usepackage[table]{xcolor}
\usepackage{booktabs}

\usepackage{enumitem}
\usepackage[font=small,labelfont=bf]{caption}
\usepackage{lipsum}
\usepackage{subcaption}
\usepackage[hidelinks]{hyperref}



\renewcommand{\thefigure}{\arabic{figure} }

\IEEEoverridecommandlockouts                              

\begin{document}

\title{\LARGE \bf
RLRC: Reinforcement Learning-based Recovery for Compressed Vision-Language-Action Models
}

\author{Yuxuan Chen$^{1}$, Yixin Han$^{1}$, Yize Huang$^{1}$, and Xiao Li$^{1}$
\thanks{$^{1}$Yuxuan Chen, Yixin Han, Yize Huang, and Xiao Li are with the State Key Laboratory of Mechanical System and Vibration, and also with the Shanghai Key Laboratory of Intelligent Robotics, School of Mechanical Engineering, Shanghai Jiao Tong University, Shanghai 200240, China
         {\tt\footnotesize \{chen\_yuxuan; hanyixin; huangyize; sjtu\_lixiao\}@sjtu.edu.cn}}%
}



\maketitle

\begin{abstract}
Vision-Language-Action models (VLA) have demonstrated remarkable capabilities and strong potential in complex robotic manipulation. However, their large parameter sizes and high inference latency hinder real-world deployment, especially on resource-constrained platforms. To address this, we conduct a systematic empirical study of model compression for VLAs. Building on these insights, we present \textit{RLRC}, a three-stage compression and recovery pipeline consisting of structured pruning, performance recovery via SFT and RL, and subsequent quantization. The RL stage incorporates a critic warm-up strategy and BC loss regularization to stabilize training and preserve policy behavior. RLRC achieves up to an 8$\times$ memory reduction and 2.3$\times$ inference speedup while maintaining the original task success rate. Extensive experiments across multiple VLA backbones show that RLRC consistently outperforms existing compression baselines, highlighting its effectiveness for on-device deployment. Project website: \href{https://rlrc-vla.github.io}{rlrc-vla.github.io}
\end{abstract}


\section{INTRODUCTION}
\IEEEPARstart{R}{ecent} advances in the field of robot learning have demonstrated new breakthroughs in both the accuracy and generalization of robotic policies for task execution. Since the introduction of RT-2 \cite{brohan2023rt}, Vision-Language-Action (VLA) models have attracted increasing attention. These models, built upon large foundation models, exhibit strong generalization capabilities. VLA models leverage the general knowledge embedded in pretrained Vision-Language Models (VLMs), while possessing the capability to comprehend language instructions, perceive the visual environment, and generate appropriate actions \cite{kim2024openvla}\cite{kim2025fine}\cite{black2410pi0}.

While VLA models hold promise for enabling general-purpose robotic capabilities, they are often derived from VLMs, which significantly limits their practicality. These models typically possess an enormous number of parameters, resulting in high computational and memory demands. Consequently, they require powerful hardware to operate efficiently, posing challenges for deployment on resource-constrained robotic platforms. Moreover, the inference speed of such models is often inadequate for real-time applications, further restricting their usability in dynamic and time-sensitive robotic tasks.
 \begin{figure}[t]
    \centering
    \includegraphics[width=0.95\linewidth]{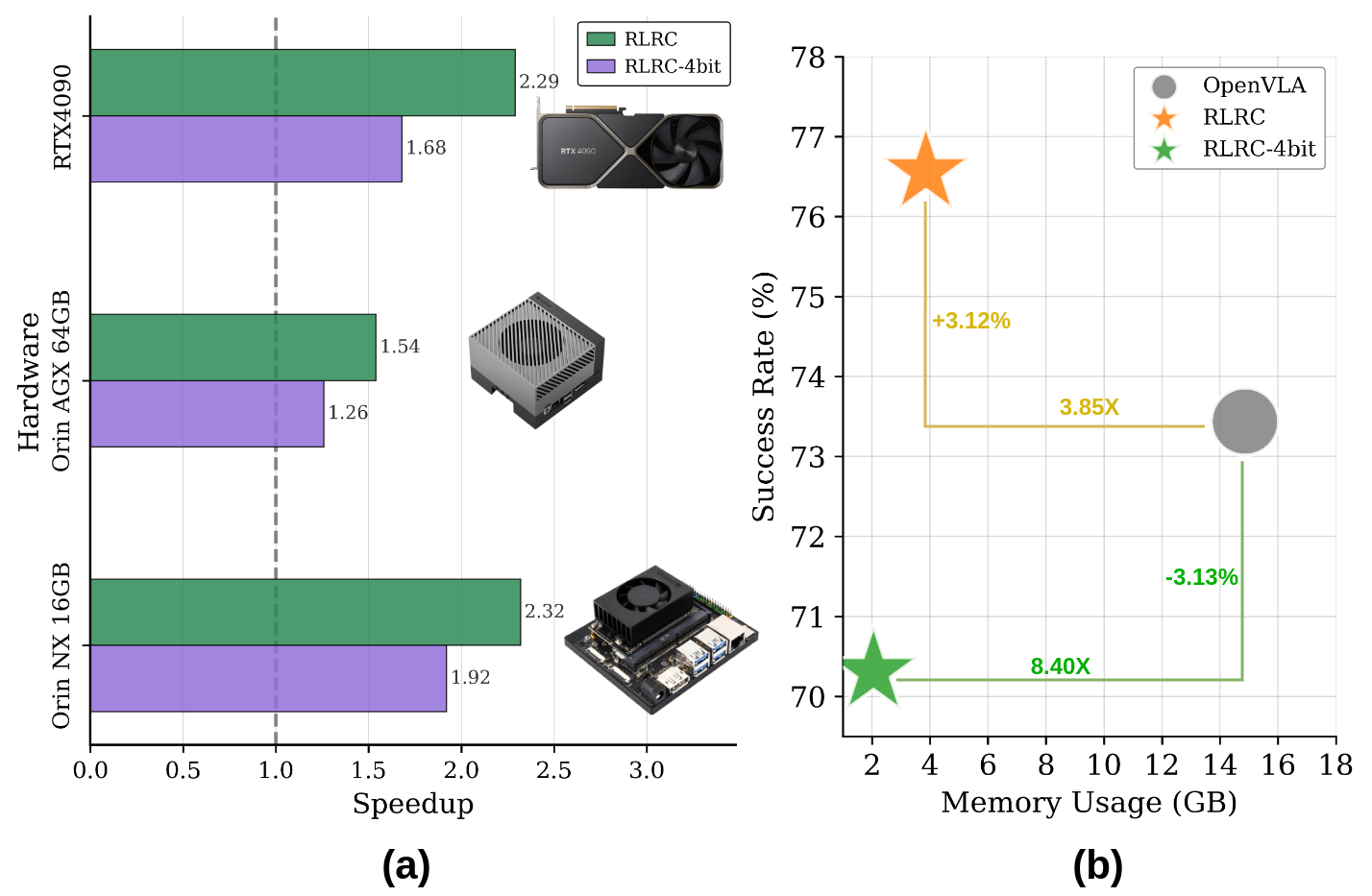}
    \caption{(a) RLRC achieves significant speedup for OpenVLA on both commonly used GPUs and Jetson edge computing platforms. (b) RLRC demonstrates notable effectiveness in preserving performance on ManiSkill tasks and reducing memory usage.}
    \label{fig:f1}
\end{figure}

To address the limitations of current vision-language-action (VLA) models, prior research has explored several strategies aimed at improving efficiency without significantly compromising performance. One prominent direction involves replacing the original backbone with more lightweight and computationally efficient architectures, thereby reducing both model size and inference latency \cite{shukor2025smolvla} \cite{Wen2024TinyVLATF}. Traditional model compression techniques, such as quantization and pruning, have been employed to shrink model size and reduce computational overhead \cite{park2024quantization}. Another effective approach involves refining visual token processing by enhancing token selection methods and implementing adaptive pruning, which helps eliminate unnecessary visual details without compromising meaningful content \cite{xu2025vla} \cite{tan2025think}. However, these methods still fall short in terms of acceleration and memory efficiency for deployment under resource-constrained settings, and they fail to adequately account for the inherent suitability of VLA for robotic control.

In this paper, we revisit this problem from a deliberately narrower perspective, conducting a systematic analysis of post-training compression and recovery for LLM-centric VLA backbones. Our study covers three representative backbones, namely OpenVLA \cite{kim2024openvla}, OpenVLA-OFT \cite{kim2025fine}, and GR00T N1.6 \cite{bjorck2025gr00t}. Our goal is practical: given an already capable but heavy VLA, derive a lighter variant through a principled pipeline that preserves most of its manipulation ability and runs faster at deployment time.

We first conduct a systematic analysis of pruning and quantization on OpenVLA. These experiments motivate a staged pipeline, RLRC, which integrates structured pruning for hardware-friendly compression, supervised fine-tuning (SFT) for initial recovery, PPO-based RL for further recovery, and optional 4-bit quantization for extreme memory reduction. It turns out that SFT repairs a large fraction of the post-pruning damage, but does not fully recover the compressed policy under the aggressive pruning regimes that deliver the largest deployment gains. In these regimes, the subsequent RL stage is not merely an extra refinement, but an empirically important part of full recovery. We further show that lower precision does not automatically imply lower latency due to dequantization overhead. As shown in Fig. \ref{fig:f1}, this staged approach leverages the strengths of each technique and offers a practical pathway for deploying VLA models on resource-constrained robotic platforms. Experimental results demonstrate that this pipeline achieves effective compression of VLA models, with minimal performance degradation, and in some cases, even surpassing the original model. The main findings and contributions of this paper are:
\begin{itemize}
    \item a staged recovery pipeline in which SFT restores the coarse behavior and PPO completes the performance recovery, together with critic warm-up and BC regularization to stabilize the SFT-to-RL transition;
    \item a systematic empirical analysis of quantization, structured pruning, and post-pruning recovery for LLM-centric VLA backbones, validated across OpenVLA, OpenVLA-OFT, and GR00T N1.6;
    \item deployment-oriented analysis covering latency--memory trade-offs, cross-backbone generalizability, and real-robot validation.
\end{itemize}

\section{RELATED WORK}\label{sec:literature review}



\subsection{Model Compression for LLMs}

LLMs have attracted widespread attention, further amplifying the demand for reducing parameter scale and lowering inference latency. To compress LLMs, there are several classic approaches: Quantization reduces memory and compute by lowering weight and activation precision, often using fewer bits\cite{frantar2023optq} \cite{liu2023llm}; pruning reduces the size and computational cost of a neural network by removing less important weights or neurons while attempting to preserve the model's performance \cite{ma2023llm} \cite{an2024fluctuation}. These model compression techniques have achieved remarkable success in the domain of LLMs, and we try to explore their application to VLA, aiming to efficiently compress VLA models while maximizing performance retention.

\subsection{Acceleration for VLAs}

Recent efforts have explored inference acceleration for VLAs. These methods differ in their reliance on additional training, as some operate in a training-free manner \cite{xu2025vla, yang2025efficientvla, jiang2025better, tan2025think} while others require fine-tuning to maintain performance \cite{jiang2025better}. A common strategy among them is to reduce computational overhead by pruning visual tokens. Although such designs improve efficiency in practice, they primarily target runtime optimization and do not fundamentally reduce the number of model parameters.

\subsection{VLA Fine-Tuning Using RL}

VLAs trained via SFT are limited by offline data, struggling with generalization due to compounding errors. To overcome this, Reinforcement Learning fine-tuning (RLFT) enables models to learn from direct environmental interaction. However, due to the inherent instability of reinforcement learning, some approaches, such as iRe-VLA \cite{guo2025improving}, adopt alternating stages of RL and supervised learning to enhance training stability. Other works, like \cite{zang2025rlinf}, introduce scalable frameworks to better handle sparse or delayed feedback. Liu et al. \cite{liu2025can} show that RL can significantly improve both semantic understanding and execution generalization compared to supervised fine-tuning alone.




\begin{table*}[htbp]
\centering
\caption{Impact of quantization and pruning on VLA performance.}
\scriptsize
\begin{tabular}{lcccccccc}
\toprule
\textbf{Model} & \textbf{spatial} & \textbf{spatial(SFT)} & \textbf{long} & \textbf{long(SFT)} & \textbf{Parameters (B)} & \textbf{Memory (GB)} & \textbf{Inference Time (ms)} & \textbf{Throughput (samples/s)} \\
\midrule
OpenVLA & \textbf{84.7} & -- & \textbf{53.7} & -- & 7.54 & 14.858 & 169 & 5.9 \\
OpenVLA + 8bit & 84.6 & -- & 52.0 & -- & 7.54 & 7.949 & 282.7 & 3.5 \\
OpenVLA + 4bit & 81.0 & -- & 49.8 & -- & 7.54 & \textbf{4.971} & \textbf{134.1} & \textbf{7.5} \\
OpenVLA + Magnitude & 83.4 & 80.4 & 51.8 & \textbf{50.6} & 7.54 & 14.826 & 162.5 & 6.2 \\
OpenVLA + Wanda \cite{sun2023simple} & 84.0 & \textbf{84.6} & 49.8 & \textbf{50.6} & 7.54 & 14.824 & 167.7 & 6.0 \\
OpenVLA + LLM-Pruner \cite{ma2023llm} & 23.4 & 84.0 & 1.0 & 46.0 & \textbf{6.23} & 12.433 & 139 & 7.2 \\
OpenVLA + FLAP \cite{an2024fluctuation} & 0.2 & 82.6 & 0.0 & 50.2 & 6.33 & 12.510 & 135.8 & 7.4 \\
\bottomrule
\end{tabular}
\label{tab:quantization and pruning}
\end{table*}

\begin{figure*}[htb]
    \centering
    \includegraphics[width=1.8\columnwidth]{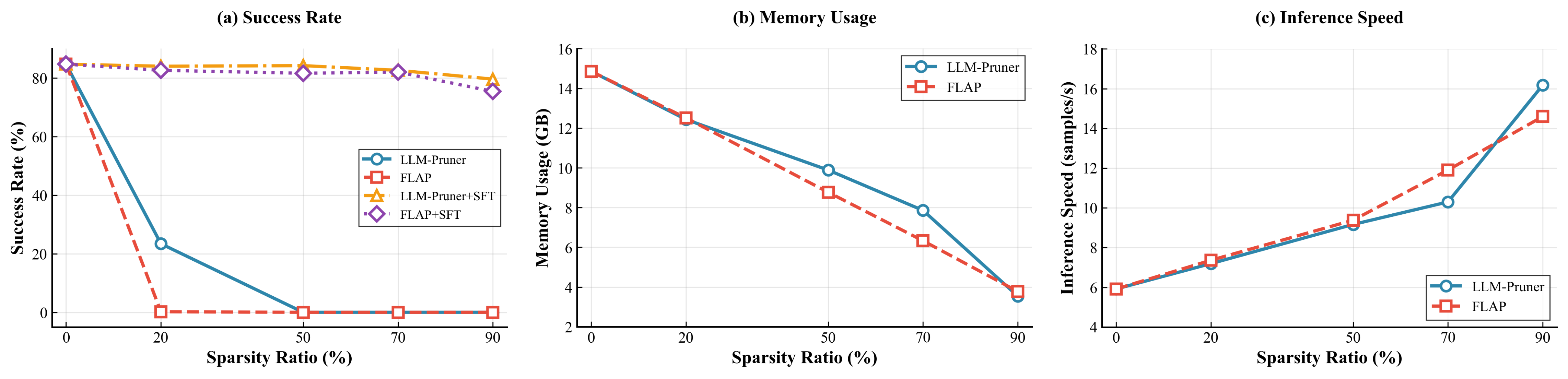}
    \caption{The impact of LLM-Pruner and FLAP on OpenVLA performance under different sparsity ratios.}
    \label{fig:prune}
\end{figure*}

\section{Empirical Study: Model Compression Applied to VLA}

We begin with a practical question: which compression tools remain effective after adapting a large VLA to a downstream robotic task? To answer this, we evaluate OpenVLA \cite{kim2024openvla} on LIBERO \cite{liu2023libero}, applying quantization and pruning exclusively to the LLM component. We use BitsAndBytes for quantization, and consider both unstructured pruning, including magnitude-based pruning and Wanda \cite{sun2023simple}, and structured pruning, including LLM-Pruner \cite{ma2023llm} and FLAP \cite{an2024fluctuation}. The results, summarized in Table \ref{tab:quantization and pruning} and Fig. \ref{fig:prune}, provide key insights that motivate RLRC.


\textbf{Quantization has minimal impact on performance, while significantly reducing memory requirements and slightly improving inference speed.} As shown in Table \ref{tab:quantization and pruning}, both 8-bit and 4-bit quantization introduce minimal degradation in the task execution performance of VLAs, while significantly reducing memory consumption. However, it is important to emphasize that quantization does not always yield improvements in inference speed. In particular, the employed LLM.int8() \cite{Dettmers2022LLMint88M} method for 8-bit quantization can incur additional computational overhead, leading to increased inference latency despite its memory efficiency.

\textbf{Unstructured pruning has a smaller performance impact, whereas structured pruning offers greater acceleration benefits.} Table \ref{tab:quantization and pruning} presents results at a 20\% sparsity ratio. Unstructured pruning has little effect on performance but provides negligible memory and inference speed gains unless combined with additional acceleration mechanisms, as it removes individual weights based on importance. In contrast, structured pruning removes entire components such as attention heads, neurons, or convolutional filters, producing regular sparsity patterns aligned with modern hardware accelerators. However, this incurs greater functional disruption by eliminating units critical to representational capacity. Consequently, structured pruning causes more severe performance degradation than unstructured pruning, as shown in Fig. \ref{fig:prune}, motivating the RLRC pipeline.

\begin{figure}[htb]
    \centering
    \includegraphics[width=0.7\columnwidth]{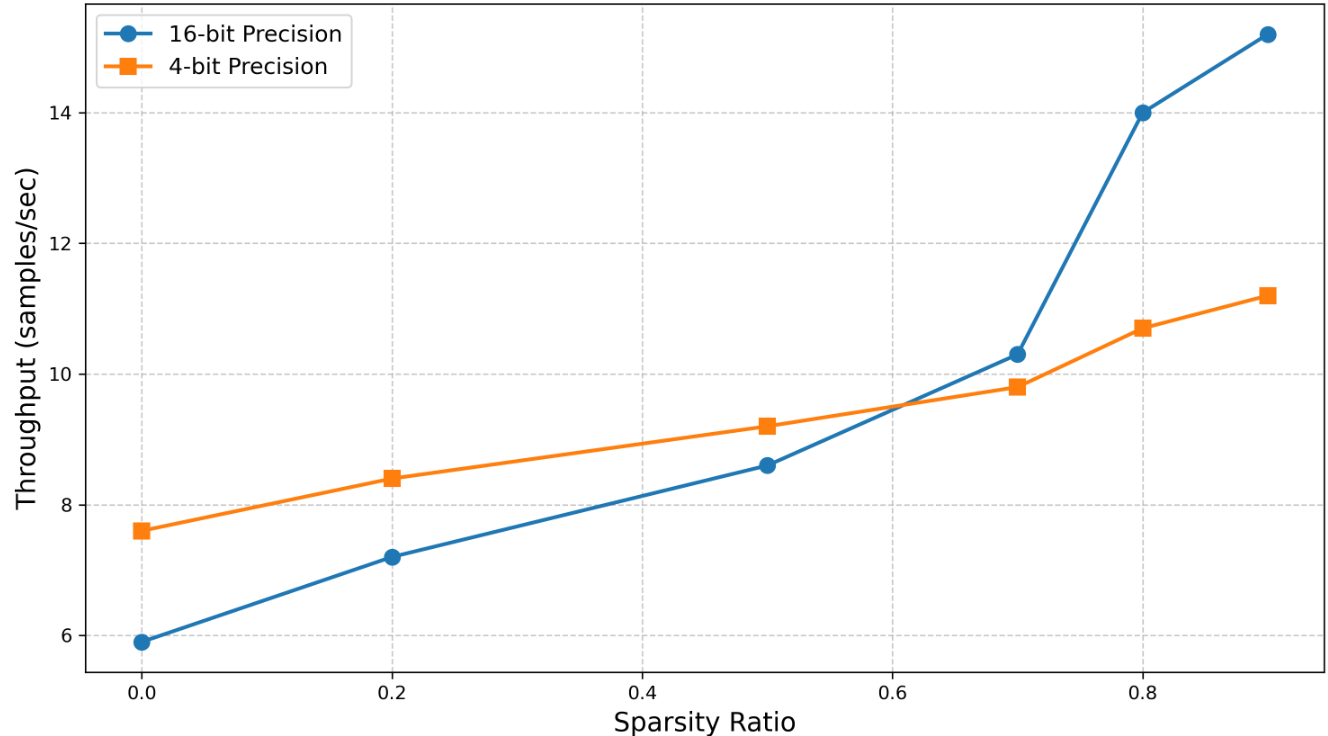}
    \caption{The impact of sparsity ratio and 4-bit quantization on OpenVLA's throughput.}
    \label{fig:speed}
\end{figure}

\textbf{The speedup gains from quantization diminish as the sparsity ratio increases.} As shown in Fig. \ref{fig:speed}, when the sparsity ratio is relatively low, quantization can significantly accelerate the inference speed of VLA. However, as the sparsity ratio increases and the model size becomes smaller, the benefits introduced by quantization are offset by the overhead of dequantization, potentially leading to slower inference. Consequently, for models with high sparsity ratios, the throughput of the full-precision version surpasses that of the quantized counterpart.

\section{RLRC}
\begin{figure*}[htb]
    \centering
    \includegraphics[width=1.9\columnwidth]{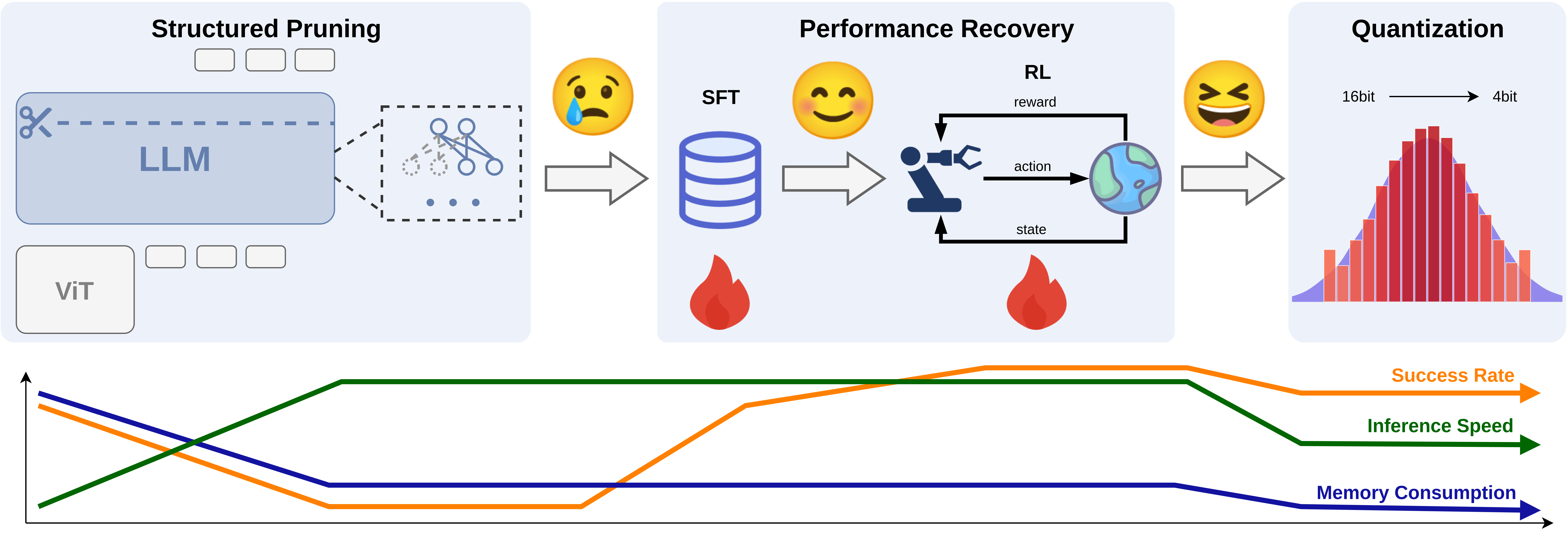}
    \caption{\textbf{Overview of the RLRC pipeline.} RLRC consists of three stages:  (1) structured pruning of VLA: structured pruning is employed with a high sparsity ratio to aggressively compress the size of the VLA; (2) performance recovery: we successfully recover the performance of the pruned VLA to a level comparable to, or even surpassing, that of the dense VLA by applying a combination of SFT and RL; (3) 4-bit quantization: we further apply 4-bit quantization to the pruned VLA, enabling deployment on resource-constrained devices at the cost of only a marginal performance degradation. The three colored curves below represent the changes in \textcolor{orange}{success rate}, \textcolor{green}{inference speed}, and \textcolor{blue}{memory consumption} throughout this process, respectively.}
    \label{fig:architecture}
\end{figure*}

Our objective is to achieve substantial reductions in model size and inference cost while preserving the task execution capabilities critical to VLA performance. To this end, we develop a three-stage pipeline, as shown in Fig. \ref{fig:architecture}: (1) we apply \textbf{structured pruning} to the VLA model, specifically targeting the LLM component, to remove redundant structures in a hardware-friendly manner (\ref{sec: sp}); (2) we employ a \textbf{performance recovery} stage that combines SFT with RL to restore the model's effectiveness on downstream tasks (\ref{sec: recover}); (3) we introduce \textbf{optional quantization} to further reduce the memory footprint, enabling efficient deployment on resource-constrained robotic platforms (\ref{sec: quantization}).
\begin{figure}[htb]
    \centering
    \includegraphics[width=0.85\columnwidth]{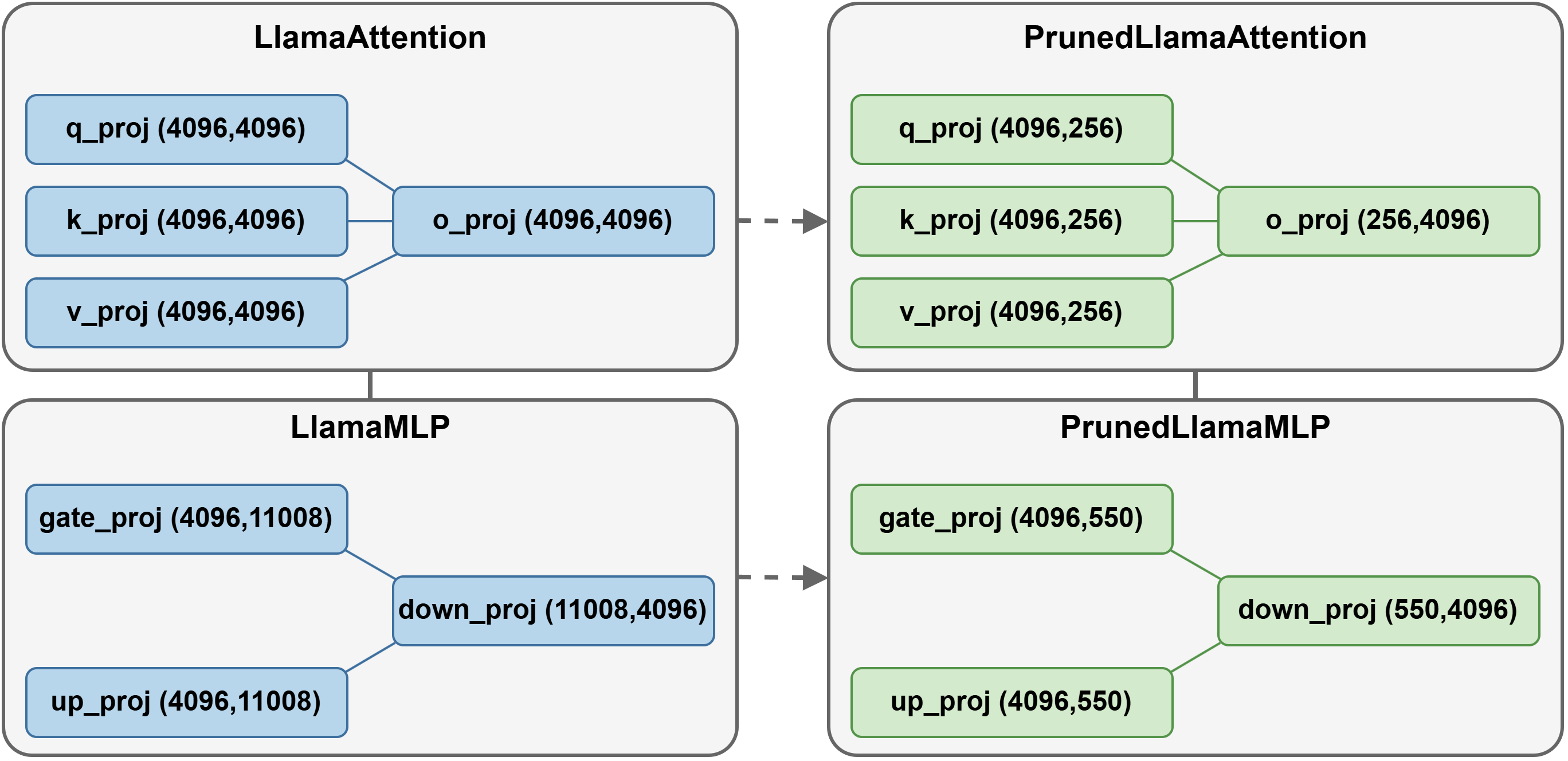}
    \caption{The attention and MLP layers within VLA before and after pruning.}
    \label{fig:vlaprune}
\end{figure}

\subsection{Structured Pruning of VLAs} \label{sec: sp} In VLAs, the majority of parameters are concentrated within the LLM component, making it a natural target for compression to reduce overall computational complexity. We specifically adopt structured pruning techniques, as opposed to unstructured pruning, due to their inherent hardware efficiency advantages, thereby enabling more substantial and practical speedups during deployment.

The first step in applying structured pruning to a VLA typically involves constructing the structural dependency graph within the LLM. Such coordinated pruning avoids introducing inconsistencies or functional degradation in the model.

We next estimate the importance of each parameter group based on its influence on the training objective. Since the pruned model is expected to undergo a subsequent recovery phase, the criterion is designed to favor groups whose removal induces minimal disruption to the training dynamics. Specifically, we incorporate the robotic expert datasets used for VLA supervised training and approximate this change in training loss using a first-order Taylor expansion. Following \cite{ma2023llm}, the importance score is formally defined as follows:
\begin{equation}
I_{g} = \left| \Delta \mathcal{L}_{g} \right| = \left| \mathcal{L}(W_{g}=0) - \mathcal{L}(W_{g}) \right| \approx \sum_{W \in g} \left| \frac{\partial \mathcal{L}}{\partial W} W \right|
\end{equation}

where $I_{g}$ denotes the importance of group $g$, $\Delta \mathcal{L}_{g}$ represents the change in the training loss evaluated on the expert dataset when the parameters $W_{g}$ belonging to group $g$ are entirely removed, and $\mathcal{L}$ indicates the objective function. 

After the definition of group importance, the pruning of the unimportant groups is executed. Formally, for a given group \( g \), we calculate its importance score \( I(g) \) and rank them based on the importance score. Then, we prune those with the lowest scores, under the assumption that they contribute least to task performance.
\begin{equation}
\mathcal{G}_{\text{pruned}} = \operatorname{Top-k}_{g \in \mathcal{G}}^{\min} \{ I(g) \}, \quad
\mathcal{G}_{\text{retained}} = \mathcal{G} \setminus \mathcal{G}_{\text{pruned}}.
\end{equation}

In all main experiments, we use a 90\% pruning ratio and keep the outermost decoder layers unchanged to preserve interface stability. Fig. \ref{fig:vlaprune} illustrates the architectural transformation of the VLA before and after structured pruning, specifically focusing on the attention and MLP layers within the LLM component.



\subsection{Performance Recovery Based on SFT and RL} \label{sec: recover}
Pruning alone creates a large performance gap, so RLRC adds two recovery stages with different roles. \textbf{SFT} is the first recovery step. We fine-tune the pruned model on the same task data using LoRA adapters and the supervised objective. This stage adapts the reduced architecture to the downstream task distribution and restores a strong initial policy. However, we observe that SFT alone is insufficient to fully recover the performance of the pruned VLA, especially when it is further subjected to aggressive 4-bit quantization, which introduces additional degradation in accuracy. To address this, we turn to Reinforcement Learning as a complementary optimization strategy.

\textbf{RL} is the second recovery step. We apply PPO \cite{schulman2017proximal} to the policy to recover the remaining performance gap, especially on harder out-of-distribution tasks. The PPO objective is:
\begin{equation}
\mathcal{L}^\theta = \mathbb{E}_t \left[
\min \left(
r_t(\theta) \hat{A}_t,\;
\text{clip}\left(r_t(\theta), 1 - \epsilon, 1 + \epsilon\right) \hat{A}_t
\right)
\right]
\end{equation}

where \( r_t(\theta) = \frac{\pi_{\theta}(a_t \mid s_t)}{\pi_{\theta_{\text{old}}}(a_t \mid s_t)} \) is the probability ratio between the new and old policy, \( \hat{A}_t \) is the estimated advantage function, \( \epsilon \) is a hyperparameter that controls the clipping range.


For RL on VLA models, the pruned VLA serves as the actor, while a lightweight value head attached to the shared transformer backbone acts as the critic. For diffusion-based VLA models, noise is injected via Flow-SDE \cite{chen2025pirl}. A sparse reward is used for all tasks: the agent incurs a constant time penalty at each step and receives a positive reward only upon task completion.

\begin{figure}[htb]
    \centering
    \includegraphics[width=0.98\columnwidth]{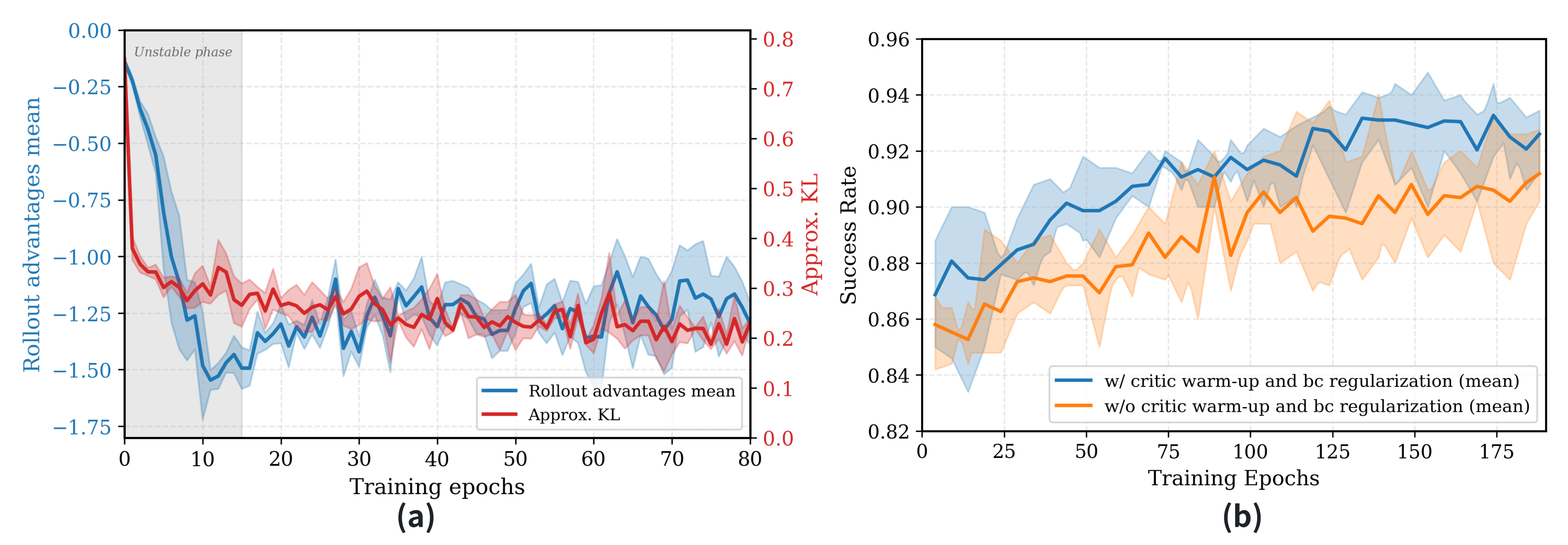}
    \caption{(a) \textbf{Early-phase training dynamics.} Curves show mean $\pm$ 1 standard deviation of rollout advantages and KL divergence over 3 runs. The shaded region marks unstable training steps; (b) \textbf{Training curves in the RL stage}, comparing configurations with critic warm-up and BC regularization versus without them. The shaded region shows the min--max range across three runs, and the solid line indicates the mean.}
    \label{fig:sft2rl}
\end{figure}

\begin{figure}[htb]
    \centering
    \includegraphics[width=0.99\columnwidth]{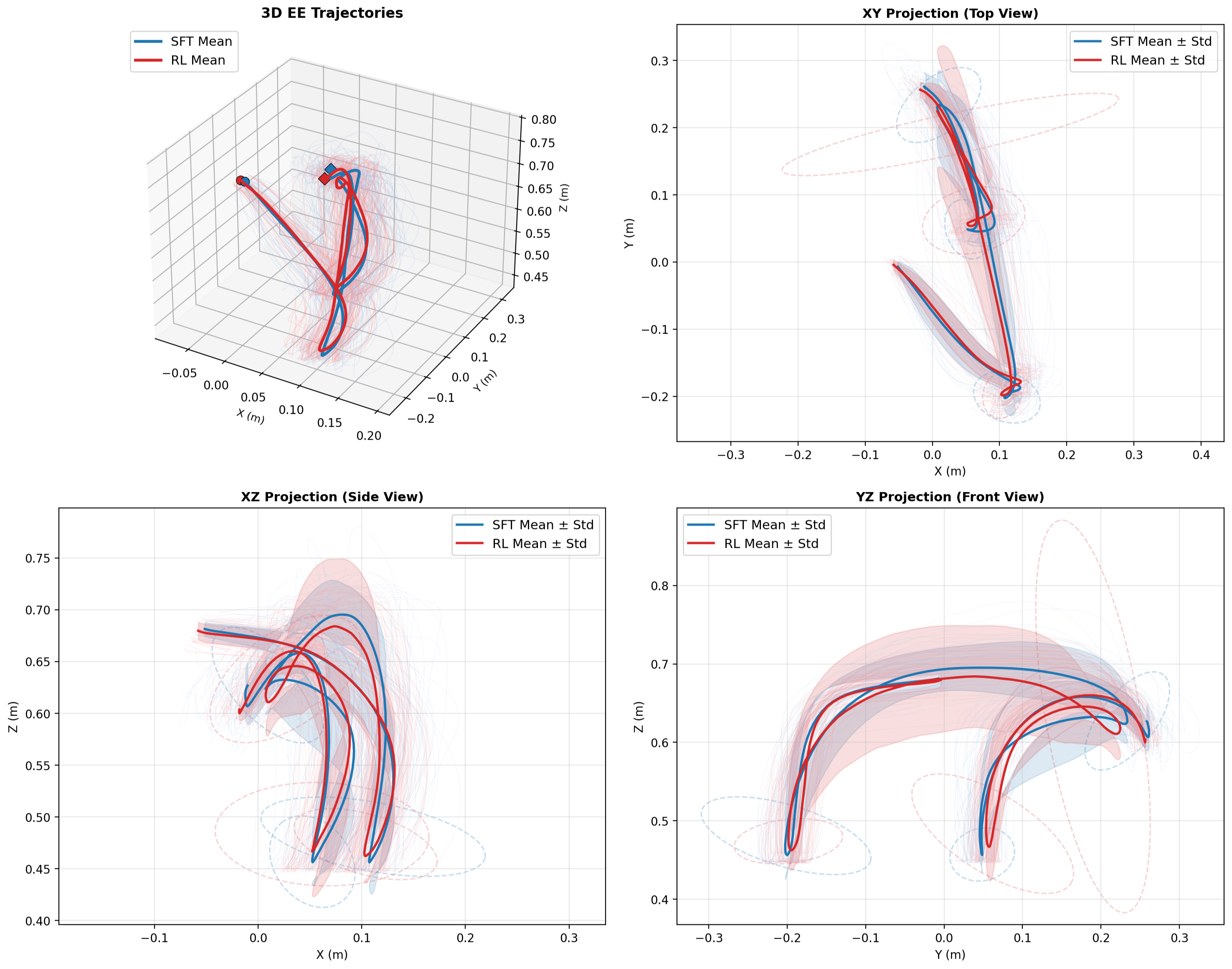}
    \caption{\textbf{End-effector trajectory comparison: SFT vs. RL rollouts.} \textbf{Top-left:} 3D Cartesian trajectories (X, Y, Z in meters); semi-transparent lines denote individual trajectories, solid lines denote means, and circles/diamonds mark start/end. \textbf{Top-right:} XY projection with mean $\pm$ 1 std bands and 2$\sigma$ covariance ellipses at 25\%, 50\%, and 75\% of trajectory length. \textbf{Bottom-left:} XZ projection. \textbf{Bottom-right:} YZ projection.}
    \label{fig:traj_vis}
\end{figure}

During training, we consistently observe a sharp performance drop when transitioning from SFT to RL. In the early RL phase, performance almost always decreases before improving, as also reported in \cite{xu2026twinrl}. This slows training and introduces instability. As shown by the orange curve in Fig. \ref{fig:sft2rl}(b), the effect is consistent across runs.

We analyze this by inspecting advantage and KL curves (Fig. \ref{fig:sft2rl}(a)). Early training exhibits unstable critic estimates that propagate to policy updates, causing erratic dynamics. Advantage signals fluctuate heavily, reducing reliability, while KL divergence varies irregularly, indicating poor update control. Comparing the SFT expert dataset with early RL rollouts (Fig. \ref{fig:traj_vis}), expert trajectories are structured and consistent, whereas rollouts show higher variance and less coherence. This mismatch indicates that the SFT-initialized policy faces a shifted data distribution during RL, hindering learning.

To address this, we modify training. First, to improve early value estimation, we introduce \textbf{critic warm-up}: freeze the VLA model and train only a randomly initialized value head, stabilizing advantage estimation and reducing policy noise. Second, to mitigate SFT--RL distribution mismatch, we add a BC loss on expert data as a regularizer, keeping the policy close to expert behavior during RL. The overall loss is:
\begin{equation}
\mathcal{L}_{\text{total}} = \mathcal{L}_{\text{PPO}} + \lambda(t) \cdot \mathcal{L}_{\text{BC}}
\end{equation}

where $\lambda(t)$ linearly decays over training. This keeps the policy close to expert behavior early on while gradually shifting to RL optimization.

These modifications significantly improve training. As shown in Fig. \ref{fig:sft2rl}(b), the early performance drop is largely reduced, enabling a smoother SFT-to-RL transition, more stable optimization, and faster gains. Final mean performance is also higher, indicating improved efficiency and policy quality.



\subsection{Further Quantization of VLAs} \label{sec: quantization}
We further explore 4-bit quantization to achieve extreme memory compression. This additional step significantly reduces the model's memory footprint, enabling seamless deployment in resource-constrained environments. Naturally, applying quantization to a VLA model pruned by 90\% further reduces memory consumption but introduces noticeable impacts on other aspects of performance. Therefore, if the target deployment device has sufficient memory, quantization may not be necessary, so we call it optional quantization and regard the variants with quantization as RLRC-4bit.


\section{Experiments}
\label{sec: experiments and results}

\begin{figure*}[htb]
    \centering
    \includegraphics[width=1.85\columnwidth]{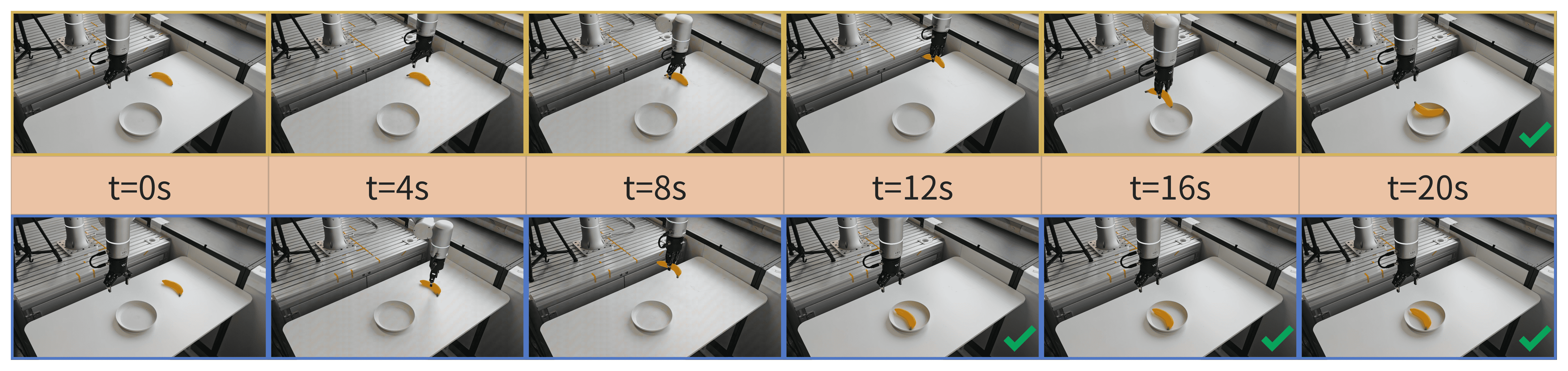}
    \caption{\textbf{Rollout examples from real-world experiments.} Video frames are sampled at equal time intervals, with OpenVLA shown on the top and OpenVLA-RLRC on the bottom.}
    \label{fig:rollout}
\end{figure*}

In our experiments, we focus on the following questions:
\begin{itemize}
    \item Q1: Is RLRC competitive compared to other methods?
    \item Q2: What is the training cost of RLRC, and how does it compare to training-free methods or inherently small policies?
    \item Q3: What is the applicability of RLRC across different VLA models?
    \item Q4: What is the impact of each component in RLRC on the overall performance?
    \item Q5: How does RLRC perform in the real world?
\end{itemize}

\subsection{Setup}
\textbf{Benchmark.} We choose LIBERO\cite{liu2023libero} and ManiSkill3\cite{tao2024maniskill3} as the benchmark simulation environments for evaluating the performance of VLAs. In ManiSkill3, we adopt the PutOnPlateInScene25Main task suite introduced by Liu et al.\cite{liu2025can}. This benchmark is divided into two splits: an in-distribution (IND) set containing 16 tasks that are seen during training, and an out-of-distribution (OOD) set comprising 9 unseen tasks. 

\textbf{Baselines.} For baseline comparisons, we apply RLRC to three representative VLA models, including OpenVLA \cite{kim2024openvla}, OpenVLA-OFT \cite{kim2025fine}, and GR00T N1.6 \cite{bjorck2025gr00t}, and obtain their corresponding RLRC variants. This design demonstrates that RLRC is compatible with diverse architectural paradigms, including both autoregressive and diffusion-based formulations. In addition to these adapted models, we include several widely used VLA approaches as baselines, such as $\pi_0$ \cite{black2410pi0}, $\pi_{0}$-Fast \cite{pertsch2025fast}, and SpatialVLA \cite{qu2025spatialvla}. To further examine performance in the low-resource regime, we incorporate a set of native lightweight models, including SmolVLA \cite{shukor2025smolvla}, UniACT \cite{zheng2025universal}, and VLA-OS \cite{gao2025vlaos}. We also compare with recent acceleration methods for VLAs, including VLA-Cache \cite{xu2025vla}, EfficientVLA \cite{yang2025efficientvla}, LightVLA \cite{jiang2025better}, and SP-VLA \cite{li2025sp}. Among these approaches, VLA-Cache, EfficientVLA, and SP-VLA operate without additional training, while LightVLA requires further optimization. 

\textbf{Metrics.} To evaluate the model’s capability in solving robotic tasks, we choose task \textit{success rate} as the primary metric. In addition to the hardware performance of VLAs, we report 3 key metrics: \textit{memory consumption}, \textit{inference time per step}, and \textit{action throughput}. 

\textbf{Implementation Details.} We implement three variants, namely OpenVLA-RLRC, OpenVLA-OFT-RLRC, and GR00T-RLRC. RL training is performed within the RLinf \cite{zang2025rlinf} framework. The 4-bit quantization is implemented based on the bitsandbytes library using the nf4 format \cite{dettmers2023qlora}. For LIBERO, success rates are computed over 500 runs per suite, and for ManiSkill, IND and OOD tasks are evaluated over 128 runs each. Hardware performance is averaged over 5 independent trials. Unless noted, all results are measured on an NVIDIA RTX 5880 Ada GPU.

\subsection{Main Results}

\begin{figure}[htb]
    \centering
    \includegraphics[width=0.8\columnwidth]{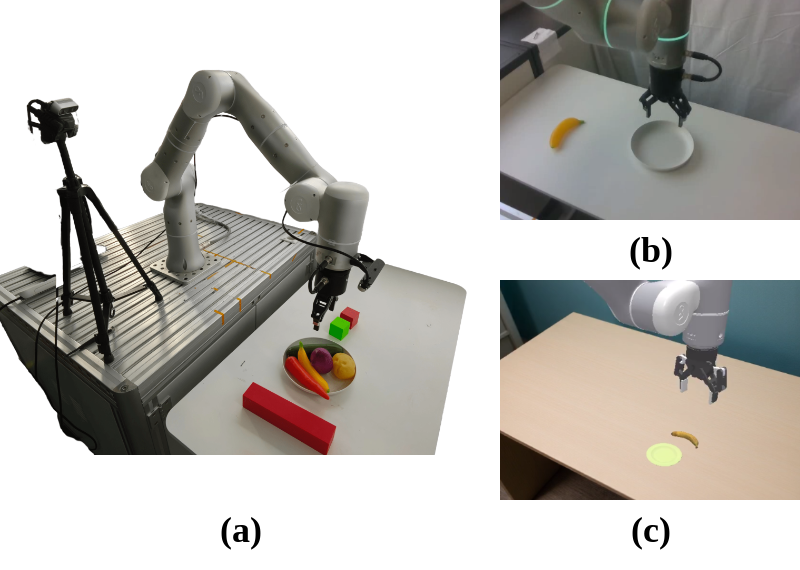}
    \caption{(a) Real-world setup with Flexiv Rizon 4. (b) Camera observation image in the real world. (c) Camera observation image in the simulator.}
    \label{fig:real_setup}
\end{figure}



\begin{table*}[t]
\centering
\caption{Comparison of different methods on LIBERO, ManiSkill, and hardware performance.}
\label{tab:vla_comparison}
\resizebox{\textwidth}{!}{
\begin{tabular}{l|ccccc|ccc|cc}
\toprule
\multirow{2}{*}{\textbf{Methods}} 
& \multicolumn{5}{c|}{\textbf{LIBERO}} 
& \multicolumn{3}{c|}{\textbf{ManiSkill}} 
& \multicolumn{2}{c}{\textbf{Hardware}} \\
\cmidrule(lr){2-6} \cmidrule(lr){7-9} \cmidrule(lr){10-11}
& \textbf{Spatial} & \textbf{Object} & \textbf{Goal} & \textbf{Long} & \textbf{Avg} 
& \textbf{IND} & \textbf{OOD} & \textbf{Avg} 
& \textbf{Model Size} & \textbf{Latency (ms)} \\
\midrule

\rowcolor{gray!15}
\multicolumn{11}{l}{\textit{Representative VLA}} \\
OpenVLA \cite{kim2024openvla} & 84.7 & 88.4 & 79.2 & 53.7 & 76.5 & 89.06 & 57.81 & 73.44 & 7B & 169 \\
OpenVLA-OFT \cite{kim2025fine} & 97.6 & 98.4 & 97.9 & 94.5 & 97.1 & 90.63 & 64.84 & 77.74 & 7B & 149.23 \\
GR00T N1.6 \cite{bjorck2025gr00t} & 98.0 & 97.8 & 98.4 & 95.4 & 97.4 & 85.16 & 53.91 & 69.54 & 3B & 75 \\
$\pi_0$ \cite{black2410pi0} & 96.8 & 98.8 & 95.8 & 85.2 & 94.1 & - & - & - & 3B & - \\
$\pi_0$-Fast \cite{pertsch2025fast} & 96.4 & 96.8 & 88.6 & 60.2 & 85.5 & - & - & - & 3B & - \\
SpatialVLA \cite{qu2025spatialvla} & 88.2 & 89.9 & 78.6 & 55.5 & 78.1 & - & - & - & 7B & - \\

\midrule
\rowcolor{gray!15}
\multicolumn{11}{l}{\textit{Native Lightweight VLA}} \\
SmolVLA \cite{shukor2025smolvla} & 93.0 & 94.0 & 91.0 & 77.0 & 88.8 & - & - & - & 2.25B & - \\
UniACT \cite{zheng2025universal} & 77.0 & 87.0 & 77.0 & 70.0 & 77.8 & - & - & - & 0.5B & - \\
VLA-OS \cite{gao2025vlaos} & 87.0 & 96.5 & 92.7 & 66.0 & 85.6 & - & - & - & 0.5B & - \\

\midrule
\rowcolor{gray!15}
\multicolumn{11}{l}{\textit{VLA Acceleration Methods}} \\
VLA-Cache \cite{xu2025vla} (OpenVLA) & 83.8 & 85.8 & 76.4 & 52.8 & 74.7 & - & - & - & 7B & 125.18 \\
EfficientVLA \cite{yang2025efficientvla} (OpenVLA-OFT) & 96.5 & 91.1 & 96.0 & 72.1 & 88.9 & - & - & - & 7B & 99.48 \\
SP-VLA \cite{li2025sp} (OpenVLA) & 75.4 & 85.6 & 84.4 & 54.2 & 74.9 & - & - & - & 7B & 124.26 \\
LightVLA \cite{jiang2025better} (OpenVLA-OFT) & 98.4 & 98.4 & 98.2 & 94.6 & 97.4 & - & - & - & 7B & 91.56 \\

\midrule
\rowcolor{gray!15}
\multicolumn{11}{l}{\textit{Ours}} \\
\textbf{OpenVLA-RLRC} & 85.2 & 88.0 & 79.6 & 52.8 & 76.4 & 92.19 & 62.50 & 77.35 & 2B & 74.04 \\
\textbf{OpenVLA-OFT-RLRC} & 97.8 & 99.6 & 98.2 & 94.8 & 97.6 & 93.75 & 68.75 & 81.25 & 2B & 65.59 \\
\textbf{GR00T-RLRC} & 98.2 & 98.0 & 97.8 & 93.4 & 96.9 & 92.19 & 66.41 & 79.3 & 2B & 55.4 \\

\bottomrule
\end{tabular}
}
\end{table*}

\textbf{RLRC demonstrates competitive performance compared to other methods (Q1).} Table \ref{tab:vla_comparison} presents the overall comparison across LIBERO, ManiSkill, and hardware metrics. RLRC consistently improves efficiency along multiple dimensions while preserving strong task performance. First, RLRC variants consistently match the performance of baseline VLAs with substantially larger parameter counts. For instance, OpenVLA-OFT-RLRC attains an average LIBERO score of 94.8, closely approaching OpenVLA-OFT's 97.4, while reducing model size from 7B to 2B parameters and latency from 149.23~ms to 65.59~ms.

At the same time, RLRC achieves notable speedup in inference. The reduced model size leads to lower computational overhead, resulting in faster execution on hardware platforms. Across different base models, RLRC demonstrates consistent latency reduction while maintaining stable performance, achieving up to 2.3× inference speedup. In terms of parameter compression, RLRC achieves up to a $3.5\times$ reduction in model size compared to the original VLA models, which translates into a comparable decrease in memory usage and contributes to substantial acceleration. Compared with existing acceleration and lightweight VLA methods, RLRC also provides a more balanced trade-off among inference speed, memory footprint, and task performance.

\textbf{Training RLRC incurs extra cost, but its benefits exceed those of training-free acceleration methods and inherently small VLAs (Q2).} RLRC cost is mainly driven by additional SFT and RL for task-specific refinement on pruned VLA models. During training, SFT converges quickly, requiring far fewer steps than training a dense VLA. The RL stage requires about 160 PPO epochs to converge, each with 128 effective episodes. Although higher than SFT, this cost remains acceptable compared to \cite{zang2025rlinf}. Moreover, aggressive pruning speeds up training compared to a dense VLA.

RLRC cost exceeds both training-free methods and inherently small policies. For example, Training RLRC on OpenVLA-OFT requires approximately 320 GPU hours, whereas fine-tuning a 2B SmolVLA requires 80 GPU hours, indicating that RLRC incurs roughly 4 times the computational cost. Despite this, RLRC yields decent gains: it maintains competitive success rates across benchmarks and consistently outperforms similarly sized small models and training-free methods by preserving strong priors from large models, enabling high-level task understanding while benefiting from parameter reduction. Compared to other methods requiring extra training, such as \cite{jiang2025better}, RLRC shows clear advantages, achieving higher task success and better generalization with a smaller memory footprint and faster inference.

\textbf{RLRC generalizes across diverse architectures, including autoregressive and diffusion-based VLAs, but has structural limitations (Q3).} Table \ref{tab:vla_comparison} reports results for OpenVLA-RLRC, OpenVLA-OFT-RLRC, and GR00T-RLRC, showing applicability to different VLA designs. This arises because RLRC operates only on the LLM, without constraining other modules. However, effectiveness varies with parameter distribution. For example, GR00T-RLRC achieves lower compression and limited acceleration, as GR00T N1.6 has fewer parameters and a smaller LLM fraction. Thus, RLRC is more advantageous for large-scale VLA models where the LLM dominates, as pruning it yields greater efficiency gains.

Despite this, RLRC has structural limitations. We observed suboptimal performance on models such as $\pi_0$ \cite{black2410pi0} and $\pi_{0.5}$ \cite{pmlr-v305-black25a}. In these models, the LLM shares transformer layers with the action expert, with tightly coupled attention and KV cache interactions. Aggressive LLM pruning disrupts the action expert's inference, causing severe degradation that is unrecoverable by SFT. Thus, RLRC suits architectures with clear separation between language and action components.

\subsection{Ablation Study}

\begin{figure}[htb]
    \centering
    \includegraphics[width=0.95\columnwidth]{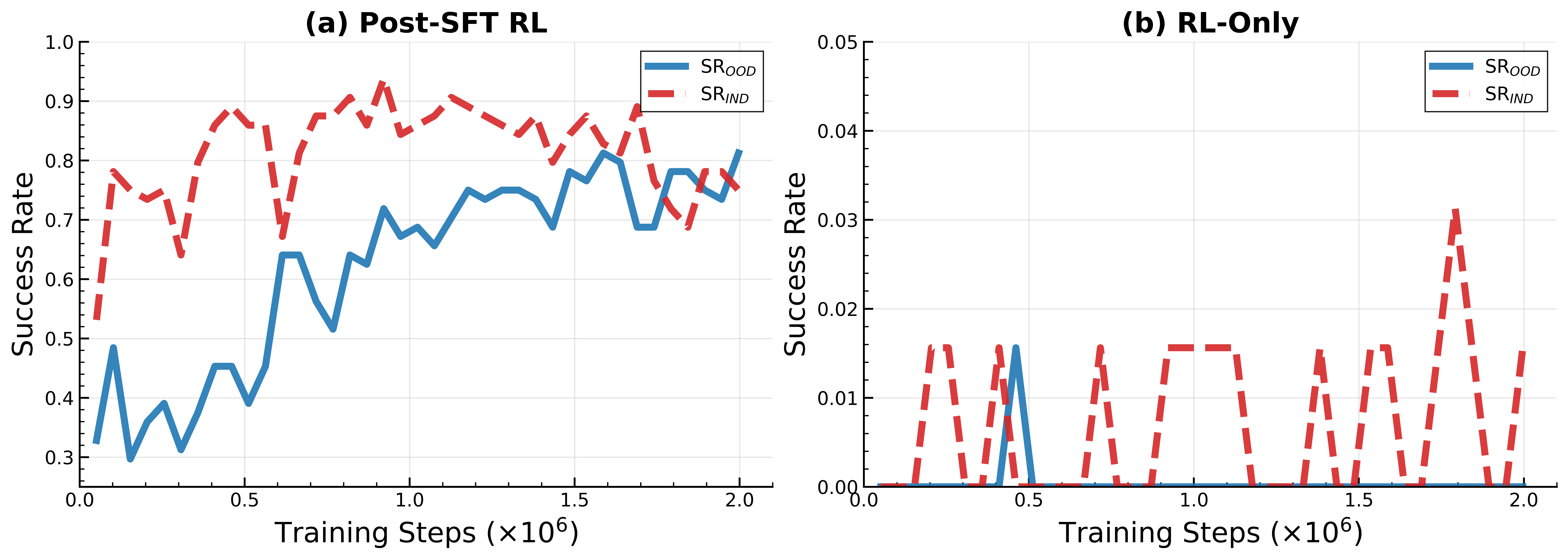}
    \caption{Training curves comparing RL after SFT with directly applying RL to the pruned VLA.}
    \label{fig:ablation_rl}
\end{figure}

\textbf{Each RLRC component plays a complementary role: SFT provides strong initialization for effective RL optimization, while critic warm-up and BC loss regularization stabilize training and improve performance (Q4).} We analyze each component via ablations, focusing on SFT, RL, critic warm-up, and BC loss regularization.

Figure \ref{fig:ablation_rl} compares training dynamics of SFT+RL with directly applying RL to the pruned model. SFT+RL yields more efficient training, which is because SFT recovers much of the original VLA performance and provides a well-initialized policy for subsequent optimization. In contrast, directly applying RL to the pruned VLA without SFT yields no meaningful gains. 

Table \ref{tab:ablation} reports success rates under different configurations, further illustrating each component's impact. We can see that in LIBERO-Long, structured pruning reduces performance from 94.5\% to 0.0\%, and SFT recovers it to 86.8\%, still significantly below the original model. In contrast, adding RL further improves performance to 94.8\%. This gap becomes more pronounced in tasks requiring long-horizon reasoning and strong generalization. So SFT alone is insufficient, especially for more complex and long-horizon tasks.

Noteably, we observe increased inference latency after quantization, likely due to extra dequantization overhead at runtime. The impact depends on model scale. For large models, memory and bandwidth reduction often outweighs dequantization, yielding overall efficiency gains. In models already heavily compressed by RLRC, dequantization cost is more pronounced since baseline computation is lightweight, making latency increase more noticeable.

\begin{table}[t]
\centering
\caption{Ablation study on LIBERO.}
\label{tab:ablation}
\scriptsize
\setlength{\tabcolsep}{4pt}
\begin{tabular}{p{2.8cm}cccc}
\toprule
\textbf{Configuration} & \textbf{Spatial} & \textbf{Long} & \textbf{Memory (GB)} & \textbf{Latency (ms)} \\
\midrule
OpenVLA-OFT & 97.6 & 94.5 & 14.98 & 149.23 \\

\hspace{2mm}$\rightarrow$ + Structured Pruning & 0.0 & 0.0 & 4.02 & \textbf{65.59} \\
\hspace{4mm}$\rightarrow$ + SFT & 92.0 & 86.8 & 4.02 & \textbf{65.59} \\
\hspace{6mm}$\rightarrow$ + RL & \textbf{97.8} & \textbf{94.8} & 4.02 & \textbf{65.59} \\

\rowcolor{gray!15}
\hspace{8mm}$\hookrightarrow$ RL w/o Critic Warm-up \& BC Reg. & 97.6 & 94.2 & 4.02 & \textbf{65.59} \\

\hspace{8mm}$\rightarrow$ + 4-bit Quant & 95.2 & 91.4 & \textbf{1.86} & 86.58 \\
\bottomrule
\end{tabular}
\end{table}

\subsection{Real-World Experiments} 

\begin{table}[t]
\centering
\caption{Results of real-world experiments. The success rate is obtained from 30 repeated trials.}
\label{tab:real-world}
\setlength{\tabcolsep}{4pt}
\renewcommand{\arraystretch}{1.15}
\begin{tabular}{lccccc}
\toprule
\textbf{Model} & \textbf{Pick} & \textbf{Stack} & \textbf{Lift} & \textbf{Memory} & \textbf{Latency} \\
 & \textbf{Place} & \textbf{Cubes} & \textbf{Peg} & \textbf{(GB)} & \textbf{(ms)} \\
\midrule
OpenVLA \cite{kim2024openvla} & 36.7 & 20.0 & 0.0 & 14.858 & 169.0 \\
\textbf{OpenVLA-RLRC} & 33.3 & 20.0 & 0.0 & \textbf{3.856} & 74.07 \\
OpenVLA-OFT \cite{kim2025fine} & \textbf{90.0} & 80.0 & 53.3 & 14.977 & 149.23 \\
\textbf{OpenVLA-OFT-RLRC} & 86.7 & \textbf{86.7} & \textbf{56.7} & 4.021 & 65.59 \\
GR00T N1.6 \cite{bjorck2025gr00t} & 83.3 & 66.7 & 40.0 & 6.765 & 75.0 \\
\textbf{GR00T-RLRC} & 83.3 & 56.7 & 43.3 & 4.269 & \textbf{55.4} \\
\bottomrule
\end{tabular}
\end{table}
\textbf{Real-World Setup.} We conduct sim2real experiments using a Flexiv Rizon 4 robot equipped with a GRAV gripper. The experiments are performed in a tabletop environment both in ManiSkill and the real world, as shown in Fig. \ref{fig:real_setup}. Real-world tasks include pick and place, stacking cubes, and lifting a peg upright. Visual observations are provided by both a third-person camera and a wrist-mounted camera, both of which are RealSense D435i sensors. OpenVLA uses a single camera for observation, while OpenVLA-OFT and GR00T use dual-camera observations together with proprioceptive state.

\textbf{RLRC demonstrates strong performance and effective compression in real-world manipulation tasks(Q5).} Due to the universal challenges of sim2real transfer, we observe that models trained in ManiSkill exhibit suboptimal real-world performance. We then collect 100 expert demonstrations through spacemouse teleoperation and use them to additionally fine-tune the models. The models are evaluated with 30 rollouts for each task. The results are summarized in Table \ref{tab:real-world}. It can be observed that RLRC still proves effective in the real world, achieving significant improvements in inference throughput and substantial reductions in memory usage, while maintaining competitive task performance. The rollout from real-world experiments is shown in Fig. \ref{fig:rollout}.

\subsection{Limitations} 
Despite its contributions, our approach presents several limitations that deserve further investigation. First, the RL component of RLRC depends on parallelized simulation environments to enable efficient learning, resulting in relatively low training efficiency and substantial computational overhead. Second, the sim-to-real gap remains a significant challenge. Policies trained in simulation may suffer from performance degradation when transferred to real-world robotic platforms due to discrepancies in dynamics, sensing, and environmental conditions.

\section{Conclusion}
\label{sec:conclusion}
In this work, we present RLRC, a systematic compression and recovery pipeline for VLAs that maintains task performance. Through a principled three-stage pipeline comprising structured pruning, performance recovery based on SFT and RL, and 4-bit quantization, we significantly reduce model size and boost inference speed while preserving, and in some cases surpassing, the original model's ability to execute robotic tasks. Our experimental results demonstrate that RLRC achieves up to
an 8× reduction in memory usage and a 2.3× increase in inference throughput, while maintaining the original VLA’s task performance. Our systematic analysis across multiple VLA backbones confirms that RLRC provides a practical and effective pathway for deploying VLAs on resource-constrained devices, advancing the real-world applicability of VLA technologies.

\bibliographystyle{IEEEtran}
\bibliography{references.bib}

\end{document}